\documentclass{article} 
\usepackage{iclr2025_conference_arxiv,times}

\usepackage{amsmath,amsfonts,bm}

\def\eqref#1{equation~\ref{#1}}

\def\1{\bm{1}}

\DeclareMathAlphabet{\mathsfit}{\encodingdefault}{\sfdefault}{m}{sl}
\SetMathAlphabet{\mathsfit}{bold}{\encodingdefault}{\sfdefault}{bx}{n}

\usepackage{hyperref}
\usepackage{url}
\usepackage{booktabs}
\usepackage{pifont}
\usepackage{caption}
\usepackage{graphicx}
\usepackage{float}
\usepackage{colortbl}
\usepackage{subcaption}
\title{CG-Bench: Clue-grounded Question Answering Benchmark for Long Video Understanding}

\author{Guo Chen$^1$\thanks{Equal contribution.}, Yicheng Liu$^1$$^*$, Yifei Huang$^2$$^*$, Yuping He$^1$, Baoqi Pei$^{2,4}$, Jilan Xu$^{2,3}$\\ \textbf{Yali Wang$^2$, Tong Lu$^1$, Limin Wang$^{1,2}$\thanks{Corresponding author.}}  \\
$^1$Nanjing University, $^2$Shanghai Artificial Intelligence Laboratory \\
$^3$Fudan University, $^4$Zhejiang University 
\\
\texttt{chenguo1177@gmail.com} \\
}

\definecolor{mygreen}{HTML}{35cd2d}

\iclrfinalcopy 
\begin{document}

\maketitle

\begin{abstract}
Most existing video understanding benchmarks for multimodal large language models (MLLMs) focus only on short videos. The limited number of benchmarks for long video understanding often rely solely on multiple-choice questions (MCQs). However, because of the inherent limitation of MCQ-based evaluation and the increasing reasoning ability of MLLMs, models can give the current answer purely by combining short video understanding with elimination, without genuinely understanding the video content.
To address this gap, we introduce CG-Bench, a novel benchmark designed for clue-grounded question answering in long videos. CG-Bench emphasizes the model's ability to retrieve relevant clues for questions, enhancing evaluation credibility. It features 1,219 manually curated videos categorized by a granular system with 14 primary categories, 171 secondary categories, and 638 tertiary categories, making it the largest benchmark for long video analysis. The benchmark includes 12,129 QA pairs in three major question types: perception, reasoning, and hallucination.
Compensating the drawbacks of pure MCQ-based evaluation, we design two novel clue-based evaluation methods: clue-grounded white box and black box evaluations, to assess whether the model generates answers based on the correct understanding of the video.
We evaluate multiple closed-source and open-source MLLMs on CG-Bench. Results indicate that current models significantly underperform in understanding long videos compared to short ones, and a significant gap exists between open-source and commercial models. We hope CG-Bench can advance the development of more trustworthy and capable MLLMs for long video understanding. All annotations and video data are released at \url{https://cg-bench.github.io/leaderboard/}.
\end{abstract}

\section{Introduction}

Recently, video understanding has made significant progress with the advent of multimodal large language models (MLLMs). To evaluate these models, many recent efforts have been made to create video understanding benchmarks~\citep{mvbench,mangalam2024egoschema,liu2024tempcompass}, providing assessments of model comprehension capabilities and clues for future improvement.

\begin{figure}
    \centering
    \includegraphics[width=\textwidth]{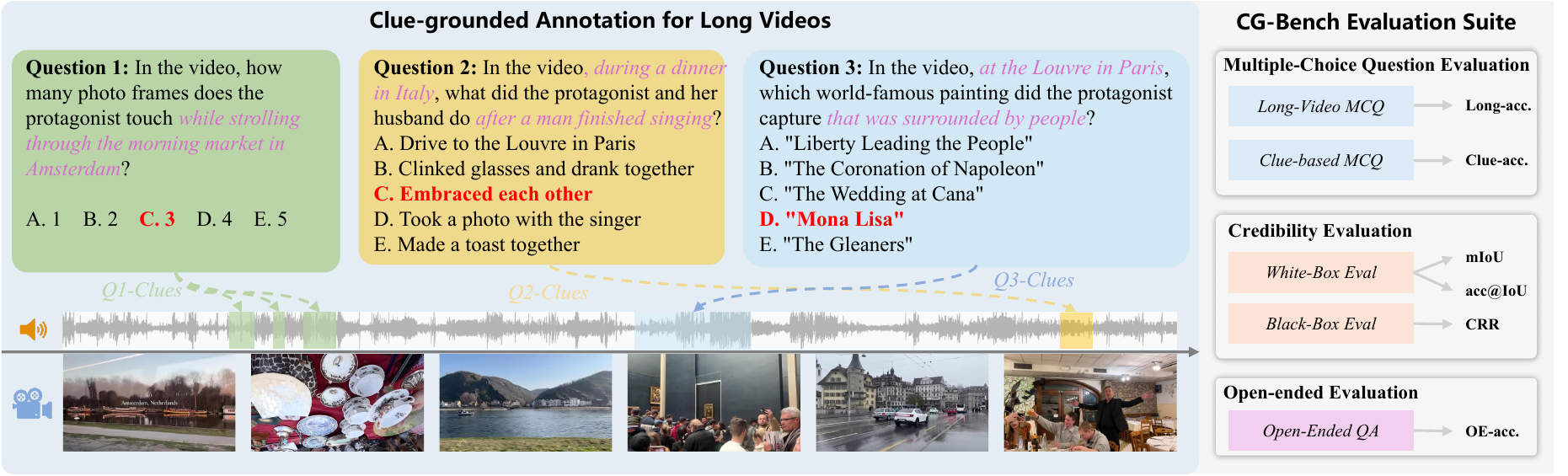}
    \caption{\textit{Left:} examples of CG-Bench's clue-grounded annotation. To correctly answer the questions, models need to ground their reasoning into the correct clue. 
    \textit{Right:} CG-Bench provides an evaluation suite with two novel credibility evaluation criteria while supporting both MCQ and open-ended evaluations.}
    \label{fig:teaser}
    \vspace{-6mm}
\end{figure}

Since early benchmarks only focus on short video clips, recent works have started to create benchmarks~\citep{videomme,wu2024longvideobench,zhou2024mlvu,huang2024egoexolearn} for longer videos ($\geq$ 10 minutes). However, these works employ multiple-choice questions (MCQ), where the difficulty level is heavily influenced by the configuration of negative options. 
In such scenarios, models~\citep{chen2023internvl,li2024llava-onevision,zhang2024llavanextvideo,lin2024vila} tend to focus on only general video knowledge and use elimination to avoid selecting the negative options. 
As a result, the models can achieve correct answers without genuinely engaging with the relevant video content, leading to a lack of trustworthiness. One illustration can be found in question 2 of Figure~\ref{fig:teaser}, the option `A' can be easily eliminated based purely on textual information.
Recently, the NExT-GQA~\citep{xiao2024nextgqa} benchmark tries to address the problem of credible models by incorporating temporal grounding into MCQ. However, NExT-GQA is limited to the NextQA~\citep{nextqa} dataset, which lacks diversity and primarily consists of short videos. 
A comprehensive benchmark for credibly evaluating \emph{generalist} MLLMs for long video understanding, is still missing in the research community.

To make up this gap, we introduce \textbf{CG-Bench}, illustrated in Figure~\ref{fig:teaser}, a novel benchmark designed to evaluate clue-grounded question answering in long videos. 
In contrast to traditional benchmarks that focus primarily on the accuracy of question answering, \textbf{CG-Bench} goes a step further by evaluating whether the model bases its answers on relevant clues within the video. 
\textbf{CG-Bench} designs two novel clue-based evaluation methods to provide more reliable model performance assessments. 1) \emph{clue-grouded white box evaluation} requires the model to directly provide the clue interval corresponding to the question while selecting the correct answer. 2) \emph{clue-grouded black box evaluation} requires the model to align the accuracy of video-level MCQ and clue-level MCQ. Furthermore, we propose a novel heuristic method, aided by human-annotated clues, for open-ended QA evaluation, to effectively balance the cost and performance.

CG-Bench features 1,219 meticulously curated videos and 12,129 human-annotated question-answer-clue (QAC) triplets, establishing it as the largest and held-out VideoQA and question grounding benchmark for long videos. It employs a highly detailed manual classification system, organizing each video into 14 primary categories, 171 secondary categories, and 638 tertiary categories. The benchmark includes three main question types: perception, reasoning, and hallucination. Perception questions are further divided into 10 subcategories, such as object and attribute recognition, while reasoning questions are categorized into 12 subcategories, including relation reasoning, etc.

We evaluate a range of closed-source and open-source MLLMs using this benchmark. The commercial models, GPT-4o \citep{gpt4o} and Gemini-1.5 Pro \citep{gemini} achieve scores of 53.9 and 43.4, respectively, with 128 frames for long-video multiple-choice questions. The leading open-source MLLM, Qwen2-VL-72B \citep{Qwen2VL}, scores 51.4 under the same conditions, indicating its initial benchmarking against GPT-4o.
However, our credibility assessments and open-ended evaluations reveal a significant drop in accuracy for existing MLLMs, with scores decreasing from 53.9 to 21.7. This underscores the considerable room for improvement in current MLLMs for long video understanding. We hope this benchmark can become a vital tool for advancing research and development of more reliable and capable MLLMs.

\section{Related Work}

\textbf{Multimodal Large Language Models (MLLMs)}
have rapidly gained popularity due to their proficiency in integrating visual and textual information~\citep{llava,llava-1.5,chen2023internvl, wang2022internvideo, wang2024internvideo2}. Recent advancements, such as LLaVA-Next-Video~\citep{zhang2024llavanextvideo}, LLaVA-OneVision~\citep{li2024llava-onevision}, and InternVL2~\citep{chen2024internvl2}, focus on enhancing MLLMs by integrating LLM backbones with visual encoders and specialized adapters, or creating higher-quality multimodal instruction data. This results in improved performance across tasks that involve both text and images.

Another area of focus is multimodal video understanding. Most models~\citep{chen2024internvl2,li2023videochat,maaz2023videochatgpt,pei2024egovideo,huang2018predicting} are optimized for short videos, typically a few seconds or at most a few minutes, without exploring their visual understanding with longer context. In response, researchers have explored methods such as compressing video frames into fewer visual tokens to allow for the handling of longer videos, as seen in models like LLaMA-Vid~\citep{llama-vid}, LVChat~\citep{wang2024lvchat}, MovieChat~\citep{song2024moviechat}, MA-LMM~\citep{he2024mallm} and Oryx~\citep{liu2024oryx}. In addition, LongVA~\citep{zhang2024longva} and LongViLA~\citep{xue2024longvila} explore the system-level optimization for long-context MLLMs which can natively support long video understanding.
Despite the continuous proposal of various MLLMs, their real-world performance in long video understanding is still under explored.

\textbf{MLLM Benchmarks.}
The development of benchmarks is becoming increasingly essential, especially for evaluating the MLLM performance in video understanding tasks. As the field develops, various benchmarks have been established to assess MLLMs across different modalities and video lengths. Previous efforts primarily focused on short videos, with traditional specialized VideoQA datasets like TVQA~\citep{tvqa}, NextQA~\citep{nextqa}, and benchmarks for MLLM like VideoBench~\citep{ning2023videobench}, MVBench~\citep{mvbench} and EgoSchema~\citep{mangalam2024egoschema}. MVBench provides a comprehensive framework for evaluating general temporal understanding capabilities through question-answering on short clips, while EgoSchema focuses on egocentric video understanding with multi-choice questions. The videos in these benchmarks typically range from a few seconds to several tens of seconds, making them similar to image benchmarks and thus hindering the development of general video LLMs.

Recently, several works such as VideoMME~\citep{videomme}, CinePile~\citep{rawal2024cinepile}, MLVU~\citep{zhou2024mlvu}, LongVideoBench~\citep{wu2024longvideobench}, MoVQA~\cite{zhang2023movqa}, and LVBench~\citep{wang2024lvbench}, have introduced long video benchmarks to evaluate MLLMs. VideoMME constructs a diverse video MCQ dataset, incorporating multimodal evaluations with visuals, subtitles, and audio. MLVU designs a range of tasks that focus on granular detail understanding to assess long video comprehension capabilities. However, a common limitation of these benchmarks is their reliance on MCQs, where the difficulty is heavily influenced by the construction of negative options. This allows MLLMs to often eliminate incorrect answers using sparse frames and common sense reasoning, which can inflate performances.
With our clue interval annotation, CG-Bench enhances the evaluation quality of MLLMs in long video understanding by introducing new evaluation mechanisms on credibility. 

\section{CG-Bench}

\subsection{Dataset Construction}

The dataset construction process of CG-Bench consists of three steps: video collection, question-answering-clue annotation, and quality review iteration. We provide details as follows.

\begin{figure*}[t]
    \centering
    \begin{minipage}[b]{0.48\textwidth}
        \centering
        \includegraphics[width=\textwidth]{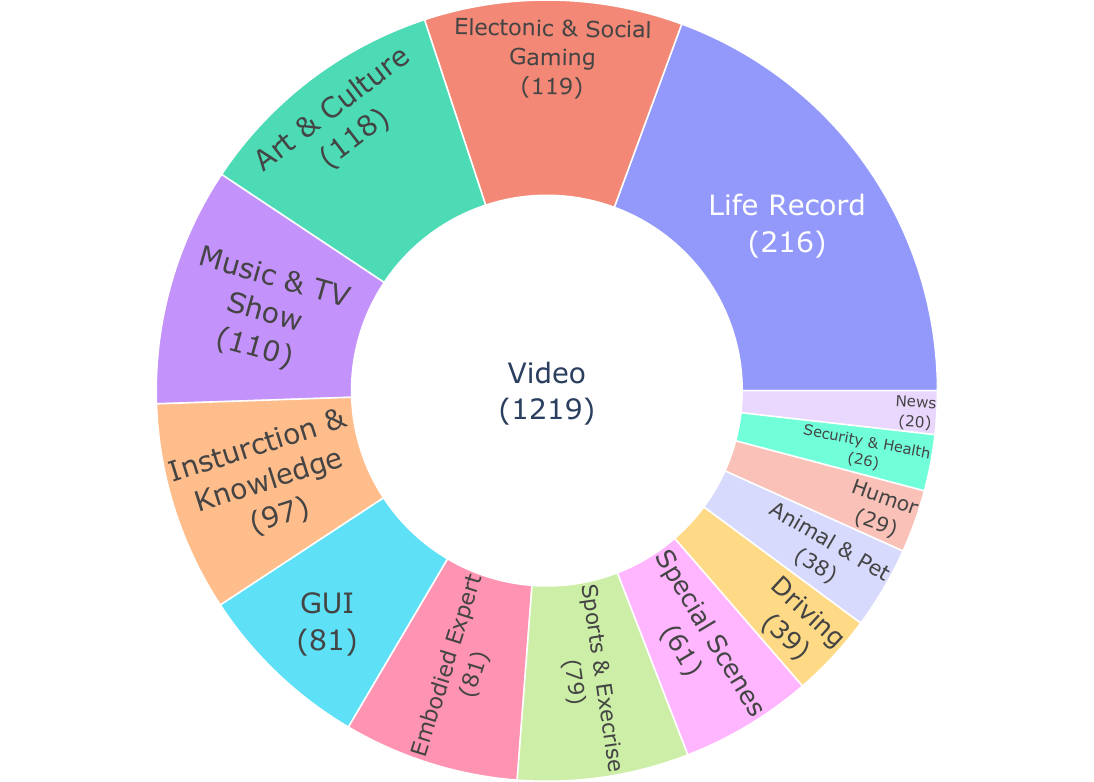}
        \caption{Distribution of video root categories, displaying the number of videos within each category.}
        \label{fig:video-categroy}
    \end{minipage}
    \hfill
    \begin{minipage}[b]{0.48\textwidth}
        \centering
        \includegraphics[width=\textwidth]{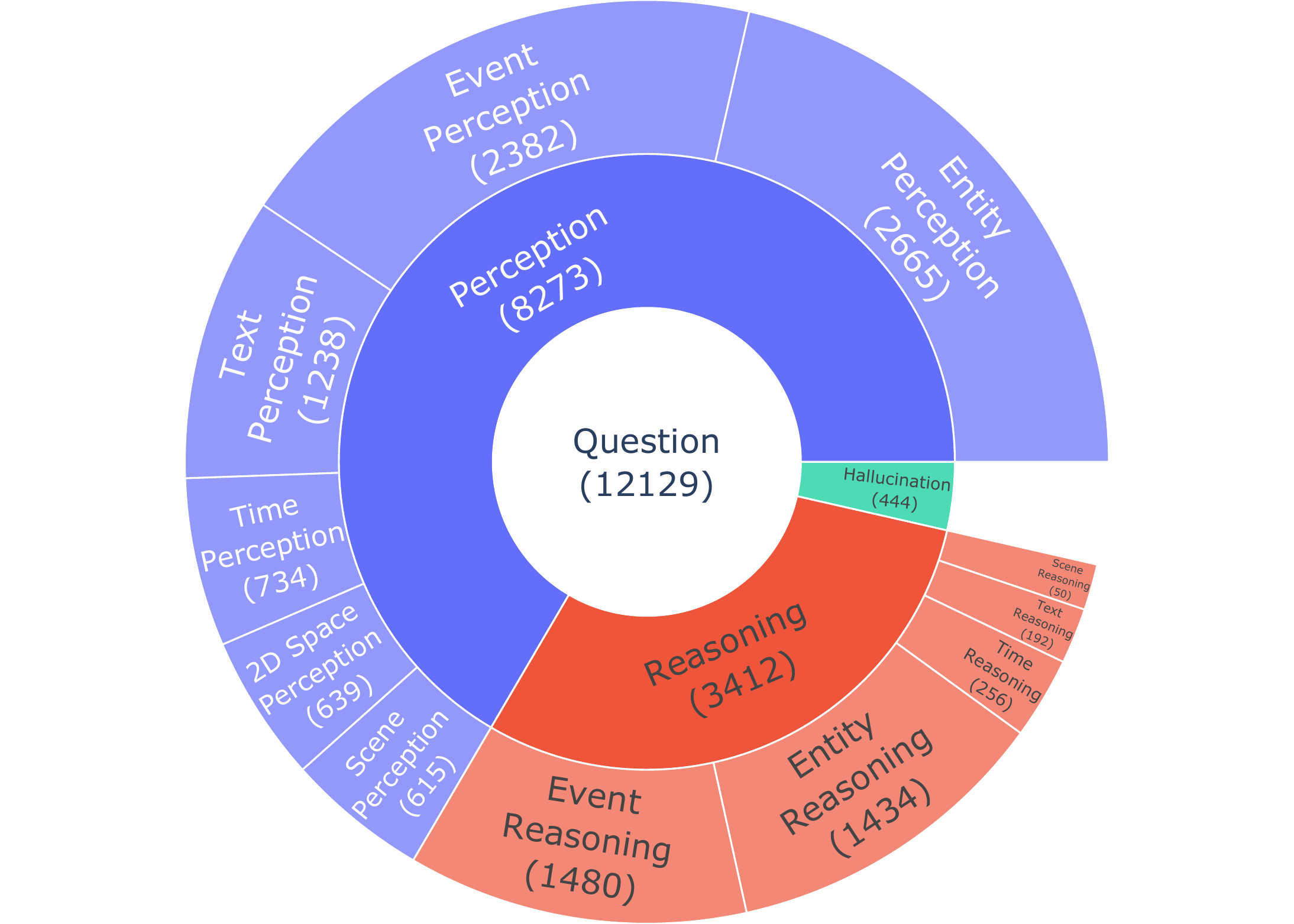}
        \caption{Distribution of question root types, illustrating the frequency of different question types.}
        \label{fig:question-category}
    \end{minipage}
    \vspace{-3.5mm}
\end{figure*}

\textbf{Video Collection.} 
To avoid using videos that have been used for pre-training by existing MLLMs, we manually collect videos from the internet and provide new annotations on them.
To facilitate the collection of raw videos from the Internet, we define 14 root domains as listed in Figure~\ref{fig:video-categroy}. 
During the collection process, we manually assign a brief tag (4-8 words) to categorize the content of each video. 
This supplementary tagging helps to ensure the diversity of the videos.
We define a video to be long if it exceeds 10 minutes in duration. 
Accordingly, we collected videos longer than 10 minutes while considering the distribution of video duration.
Furthermore, we retain the accompanying subtitles and audio to provide multimodal information. 
We carefully review and filter the videos manually for 7 rounds. 
More details about the video collection can be found in the supplementary material.

\textbf{Question-Answer-Clues Annotation.} After collecting the raw video data, we annotate it with high-quality question-answer-clue (QAC) triplets. To ensure question diversity, we first establish a taxonomy with three main types: Perception, Reasoning, and Hallucination. As shown in Figure~\ref{fig:question-category}, Perception and Reasoning questions are further divided into 10 and 14 subcategories, respectively, while Hallucination questions combine elements of both perception and reasoning.
Annotators are instructed to include negative options to create a multiple-choice QA format, facilitating straightforward and cost-effective assessments. To minimize expression loss, annotators use their native language during the annotation process. Each video is annotated 6 to 15 QAC triplets depending on its duration.
To ensure consistency in QAC triplets, we standardized the annotation process by first annotating the QA pairs and then identifying the clues. Annotators must watch the entire video, select a question type from the predefined categories, and then annotate a new question and its corresponding answer. Next, they select one or more intervals from the video to form a QAC triplet. Since the actual clue intervals often consist of multiple short moments, annotating each fragment is costly. Therefore, annotators are required to mark intervals that cover these short moments while ensuring the completeness of each event.

\textbf{Review Iteration.} 
To ensure the difficulty and quality of the dataset, we conduct a repetitive review and iteration process to enhance annotation quality. 
We reject annotations that do not meet our quality standards and request annotators to revise them. Our quality requirements for annotations and the measures taken to ensure them are as follows:
1) \emph{The rationality of the question, options, and answer}: we conduct manual reviews;
2) \emph{The video dependency of the question, options, and answer}: we input questions and options into GPT-4 and filter out QA pairs that can be answered solely based on pure text;
3) \emph{The difficulty of negative options in multiple-choice questions}: we input the video, questions and options into MLLMs and filter out QA pairs that can be answered using only sparse frames and small models;
4) \emph{The positional diversity of clue intervals:} We monitor the distribution of clue duration and position and provide timely guidance to annotators.

\subsection{Dataset Statistics \& Comparisons}

We present the detailed statistics of our dataset to provide a more comprehensive understanding, including meta-information, QAC triplets, qualitative analysis, and comparison to previous works.

\subsubsection{Dataset Statistics} 
\begin{figure*}[t]
    \centering
    \begin{minipage}[b]{0.32\textwidth}
        \centering
        \includegraphics[width=\textwidth]{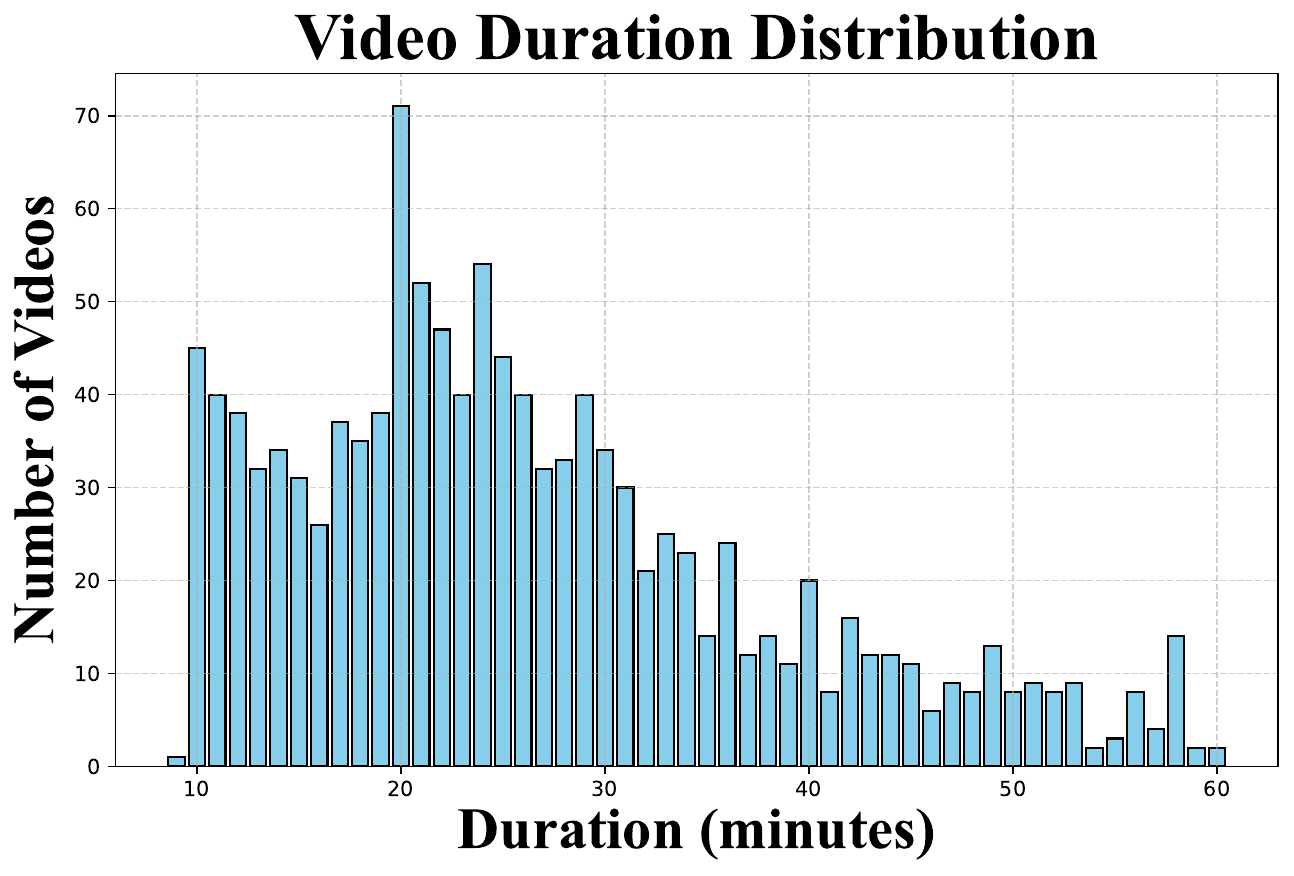}
        \caption{Video duration distribution, showing the number of videos for different duration intervals.}
        \label{fig:video-duration-dis}
    \end{minipage}
    \hfill
    \begin{minipage}[b]{0.32\textwidth}
        \centering
        \includegraphics[width=\textwidth]{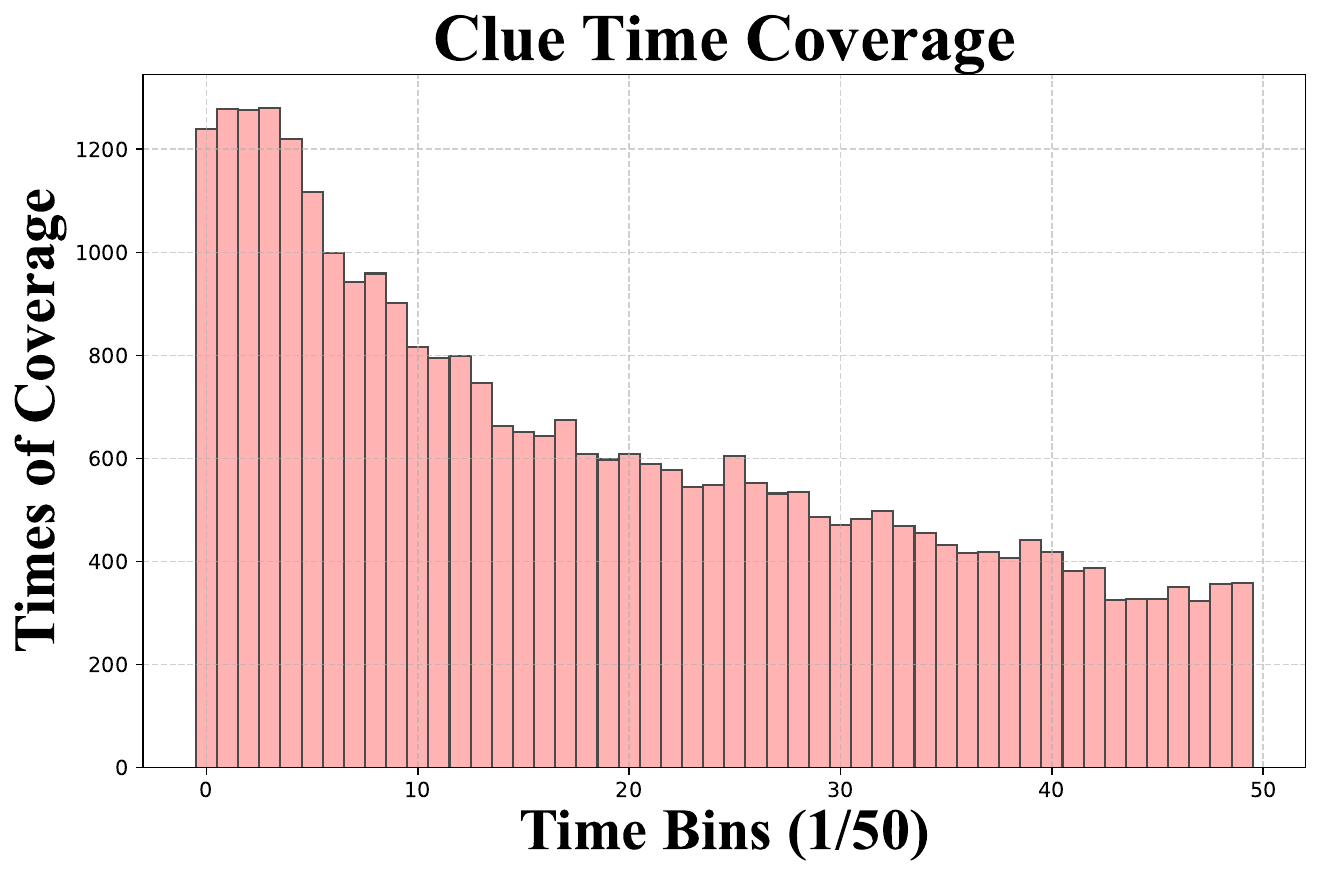}
        \caption{Clue time coverage, illustrating the frequency of clues across different time bins.}
        \label{fig:clue-time-dis}
    \end{minipage}
    \hfill
    \begin{minipage}[b]{0.32\textwidth}
        \centering
        \scriptsize
        \begin{tabular}{ll}
            \toprule
            Annotation Statistics \\
            \midrule
            \#QAC Triplets & 12129  \\
            \#Avg/QAC per video & 9.95 \\
            \#Avg/Option per QAC & 6.96  \\
            \#Avg/Clue per QAC & 1.18  \\ 
            \midrule
            \#Avg/Words of Questions & 20.07  \\
            \#Avg/Words of Options & 22.88  \\
            \#Avg/Duration of Clues & 19.24  \\
            \bottomrule
        \end{tabular}
        \captionof{table}{Annotation statistics, detailing the number of QAC triplets, questions, options, and clues.}
        \label{tab:annotation-statistics}
    \end{minipage}
    \vspace{-5mm}
\end{figure*}

\textbf{Video Meta.} Our dataset comprises a total of 1219 videos with multimodal information, including vision, audio, and subtitles. The duration of the videos varies between 10 and 80 minutes, with a distribution illustrated in Figure~\ref{fig:video-duration-dis}. Notably, videos that last between 20 and 30 minutes are the most prevalent. This selection process is manual, based on content relevance, which mirrors real-world duration distributions and highlights a long-tail effect for longer videos.
As illustrated in Figure~\ref{fig:video-categroy}, each video is classified using a three-tiered tagging system that succinctly encapsulates its content and assigns it to fundamental categories. The primary classification is augmented by a secondary layer of 171 tags and a tertiary layer consisting of 638 tags. This multi-level tagging mechanism guarantees a broad diversity of data content. For a more detailed classification of tags, please consult the supplementary materials.

\textbf{QAC Annotation.} CG-Bench includes 12,129 annotations consisting of questions, answers, and clues. Table~\ref{tab:annotation-statistics} presents the sentence lengths and totals for the annotated questions and answers, highlighting the linguistic diversity within our dataset.
Each QAC triplet is annotated with 4 to 7 negative samples, resulting in an approximately uniform distribution with ratios of options A to H of 12.4\%, 14.7\%, 12.1\%, 14.8\%, 15.1\%, 16.1\%, 11.6\%, and 3.1\%. 
There are a total of 14,362 clue intervals across all QAC triplets, with an average duration of 19.24 seconds each. Additionally, we conduct a further analysis of the positions of clue intervals within the video. Figure~\ref{fig:clue-time-dis} illustrates the frequency with which each normalized timestamp is represented by intervals. This demonstrates the unbiased nature of our interval annotations and highlights the diversity of our QA content in temporal position.

\begin{table}[t]
\centering
\scriptsize
\caption{Comparison of benchmarks across key aspects: number of videos (\#Video), average duration (\#Duration), number of QA pairs (\#QA Pairs), number of clues (\#Clue), annotation method (M/A for manual/automatic), Open-Domain (OD), Open-Ended (OE), Multi-modal (MME), and Credibility (CE) Evaluation.}
\label{comparison-bebchmark-table}
\setlength{\tabcolsep}{2mm}{
\begin{tabular}{lccccccccc}
\toprule
\textbf{Benchmark}  &\textbf{\#Video} &\textbf{\#Dur.(s)} &\textbf{\#QA Pairs} &\textbf{\#Clue} &\textbf{Anno.} & \textbf{OD}  & \textbf{OE} & \textbf{MME} & \textbf{CE} \\ 
\midrule
\emph{Question-Clue Grounding} \\
NextGQA~\citep{xiao2024nextgqa}        &1,000 & 39.5 & - & 10,531 & M & \ding{55} & - & - & -\\
Ego4D-NLQ$_\texttt{val}$~\citep{ego4d} &415 & 499.7  & - & 4,554 & M & \ding{55} & - & - & -\\
Ego4D-NLQ$_\texttt{test}$~\citep{ego4d} &333 & 493.7  & - & 4,005 & M & \ding{55} & - & - & -\\
MultiHop-EgoQA$_\texttt{test}$~\citep{chen2024multihop-egoqa} & 360 & -  & - & 1,080 & A\&M & \ding{55} & - & - & -\\
E.T. Bench$_\texttt{test}$~\citep{liu2024etbench} & - & 129.3  & - & 2,011 & M & \checkmark & - & - & -\\
RexTime$_\texttt{test}$~\citep{chen2024rextime} &- & 141.1  & - & 2,143 & A\&M & \ding{55} & - & - & -\\
\rowcolor{pink!50}\textbf{CG-Bench-QG}             &1,219 & 1624.4  & - & 14,362 & M & \checkmark & - & - & -\\

\midrule
\emph{Short-Video QA} \\
TVQA~\citep{tvqa} & 2,179 & 11.2 & 15,253 & 15,253 & M & \ding{55} & \ding{55} & \ding{55} & \ding{55} \\
STAR~\citep{star} & 914 & 11.9 & 7,098 & 7,098 & A & \ding{55} & \ding{55} & \ding{55} & \ding{55} \\
NextQA~\citep{nextqa} & 1,000 & 44.0 & 8,564 & \ding{55} & A & \ding{55} & \checkmark & \ding{55} & \ding{55} \\
EgoSchema~\citep{mangalam2024egoschema} & 5,063 & 180.0 & 5,063 & \ding{55} & A\&M & \ding{55} & \ding{55} & \ding{55} & \ding{55} \\
TempCompass~\citep{liu2024tempcompass} & 410 & 11.4 & 7,540 & \ding{55} & A\&M & \ding{55} & \ding{55} & \ding{55} & \ding{55} \\
RexTime$_\texttt{test}$~\citep{chen2024rextime} &- & 141.1  & - & 2,143 & A\&M & \ding{55} & \ding{55} & \ding{55} & \checkmark\\
MVBench~\citep{mvbench}         & 3,641 & 16.0 & 4,000 & \ding{55} & A\&M & \ding{55} & \ding{55} & \ding{55} & \ding{55} \\
MMBench-Video~\citep{fang2024mmbenchvideo}             & 600 & 165.4 & 1,998 & \ding{55} & M & \checkmark & \checkmark & \ding{55} & \ding{55} \\
\rowcolor{pink!50}\textbf{CG-Bench-Clue}          & 12,129 & 22.8  & 12,129 & - & M & \checkmark & - & \checkmark & - \\

\midrule
\emph{Long-Video QA} \\
EgoTimeQA$_\texttt{test}$~\citep{di2024groundedvqa} & 148 & 492  & 500 & \ding{55} & A & \ding{55} & \ding{55} & \ding{55} & \ding{55} \\
MovieChat-1K~\citep{song2024moviechat} & 130 & 500.0 & 1,950 & \ding{55} & M & \ding{55} & \ding{55} &\ding{55} &\ding{55} \\
Video-MME~\citep{videomme} & 900 & 1017.9  & 2,700 & \ding{55} & M & \checkmark & \ding{55} & \checkmark & \ding{55} \\
LongVideoBench~\citep{wu2024longvideobench} & 966  & 1408.0  & 6,678 & \ding{55} & M & \checkmark&  \ding{55} & \ding{55} & \ding{55}\\
MLVU~\citep{zhou2024mlvu} & 757  & 720.0 & 2,593 & \ding{55} & M & \ding{55} & \ding{55} & \ding{55} & \ding{55} \\

\rowcolor{pink!50}\textbf{CG-Bench}             & 1,219  & 1624.4 & 12,129 & 14,362 & M & \checkmark & \checkmark & \checkmark & \checkmark \\

\bottomrule
\end{tabular}
}
\vspace{-5mm}
\end{table}

\subsubsection{Comparison with Previous Benchmarks} CG-Bench is characterized by its diverse features, allowing it to be compared with three distinct types of benchmarks, as depicted in the three sections of Table~\ref{comparison-bebchmark-table}: Question Clue Grounding, Short-Video QA, and Long-Video QA benchmarks. For the question clue grounding benchmarks, NextGQA~\citep{xiao2024nextgqa}, Ego4D-NLQ~\citep{ego4d}, MultiHop-EgoQA~\citep{chen2024multihop-egoqa}, E.T. Bench~\citep{liu2024etbench}, and RexTime~\citep{chen2024rextime} are primarily centered around action and egocentric domains. Their videos are sampled from academic datasets.
In comparison, the question clue grounding part of CG-Bench, CG-Bench-QG, stands out with the highest number of videos and the longest average length, the diversity of which fosters a broad spectrum of question-grounding queries. 

Furthermore, we transform QAC triplets to our novel Short-Video QA benchmark, termed CG-Bench-Clue. When contrasted with prior short video benchmarks such as TempCompass~\citep{liu2024tempcompass}, MVBench~\citep{mvbench} and MMBench-Video~\citep{fang2024mmbenchvideo}, our CG-Bench-Clue emerges as the \emph{\textbf{largest}}, \emph{\textbf{held-out}}, \emph{\textbf{open-domain}} and \emph{\textbf{multimodal}} Short-Video QA benchmark. 

As for the Long-Video QA benchmark, CG-Bench excels in the number of videos, length, quantity of questions, and annotation quality. Owing to our clue interval annotations, CG-Bench further facilitates reliable evaluations for long videos and open-ended evaluations with clue assistance, a feature that sets it apart from existing long video benchmarks like Video-MME~\citep{videomme} and MLVU~\citep{zhou2024mlvu}.

\subsection{Evaluation}
In this section, we describe the evaluation tasks of our CG-Bench which include traditional MCQ, the unique credibility evaluation, and clue-aided open-ended QA evaluation.

\subsubsection{Multiple-Choice Question Evaluation}
We assess the accuracy of MCQ in two settings: \textbf{Long-Video MCQ} and \textbf{Clue-based MCQ}. In the Long-Video MCQ setting, the model receives the entire video as input and is required to select the correct answer based on the video, the question, and the candidate options. For the Clue-based MCQ setting, the model is given only the video within the annotated clue interval as input. The model has access only to the clue clip, the question, and the candidate options. It does not have access to the original long video. Since a single QA may correspond to multiple clues, we merge these clues and treat the combined clue as a single, cohesive clue segment.

\subsubsection{Credibility Evaluation}
The ability of a model to identify relevant clues related to questions is a crucial factor in determining its reliability. Therefore, we define a model's reliability based on its proficiency in locating accurate clues when addressing problems. To achieve this, we introduce two clue-grounded mechanisms for credibility assessment: white-box evaluation and black-box evaluation.

\textbf{White-Box Evaluation} requires the model to directly output the intervals of clues that can accurately answer the question. This task is similar to video temporal grounding~\citep{moment_detr_qvhighlights,huang2023weakly}. Therefore, we use tIoU (Temporal Intersection over Union) as the evaluation metric. Since each question may correspond to multiple intervals of clues, we allow the model to predict multiple possible intervals. Given a set of prediction \textbf{$\mathcal{P}$} and ground truths $\mathcal{G}$, the tIoU is defined as:
\begin{equation}
\small
\text{tIoU} = \frac{\sum_{i \in \mathcal{G}, j \in \mathcal{P}} \max(0, \min(b_i, d_j) - \max(a_i, c_j))}{\sum_{i \in \mathcal{G}} (b_i - a_i) + \sum_{j \in \mathcal{P}} (d_j - c_j) - \sum_{i \in \mathcal{G}, j \in \mathcal{P}} \max(0, \min(b_i, d_j) - \max(a_i, c_j))} \times 100\%,
\end{equation}
where $a_i$, $b_i$ are the start and end timestamps of the $i$-th ground truth interval of $\mathcal{G}$. $c_j$, $d_j$ are the start and end timestamps of the $j$-th predicted interval of $\mathcal{P}$. 
We calculate the mean IoU (\textbf{mIoU}) by averaging the tIoU scores obtained by the model across all question queries. To further improve the robustness of question grounding evaluation, we introduce the \textbf{rec.@IoU} metric. This metric measures the probability of successfully recalling clue intervals at various IoU thresholds. We calculate the average recall rate at IoU thresholds of 0.1, 0.2, 0.3, 0.4, and 0.5 to determine the final result. 

In addition, we propose a combined metric, \textbf{acc.@IoU} that evaluates both MCQ accuracy and clue-grounding ability. For a question with multiple choice options, the response is considered correct only if the selected answer is accurate and the tIoU between the prediction and the ground truth exceeds ($>$) a predefined threshold $\tau$. Since locating short-duration clues in the long videos in CG-Bench is inherently challenging, we set the default $\tau$ to be 0 for the more obvious comparison on ablation studies. 
Setting $\tau=0$ ensures that acc.@IoU requires the model to select the correct option and produce a time interval that overlaps at least slightly ($\text{tIoU}>0$) with the annotated clue interval, rather than reducing to naive MCQ accuracy. We calculate the \textbf{acc.@IoU} at IoU thresholds of 0.1, 0.2, 0.3, 0.4, and 0.5 to determine the final result.

\textbf{Black-Box Evaluation} aims to evaluate the model's ability to seek out clues implicitly. Understanding long videos involves the retrieval of clues distributed across various spatiotemporal locations within the entire video. Therefore, an effective model for long videos should naturally focus on capturing human-annotated clue intervals in its hidden states. However, beyond the explicitly annotated clue intervals, there are likely hidden clues scattered throughout the video that can also help to determine the correct answer.
Thus, a model with access to the full video should yield higher accuracy compared to solely relying on the clue interval. In other words, the accuracy of Long-Video MCQ ( \textbf{long-acc.}) should be greater than or equal to the accuracy of Clue-based MCQ (\textbf{clue-acc.}).

With this insight, for the black box evaluation, we define a new metric called Clue Recovery Rate (\textbf{CRR}).  
This metric evaluates the model's robustness to context dilution, \textit{i.e.}, how stable a model can find related clues from long but diluted video context.
CRR is calculated by:
\begin{equation}
\small
\text{CRR} = \frac{\min(\textbf{long-acc.}, \textbf{clue-acc.})}{\textbf{clue-acc.}} \times 100\%,
\end{equation}
A CRR of less than 100\% suggests that the MLLM's ability to retrieve short clues from long video representations is not optimal.

\begin{figure}[t]
    \centering
    \includegraphics[width=\textwidth]{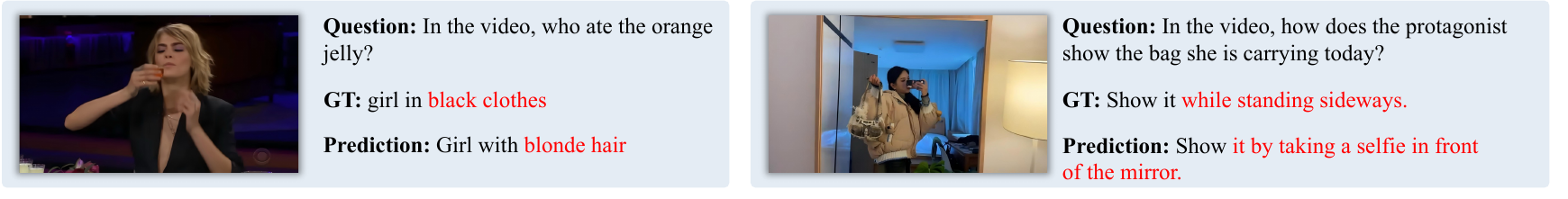}
    \caption{Two examples illustrating the ambiguity challenge of using LLMs for open-ended evaluation. While in different expressions, GT and prediction should both be treated as correct answers.}
    \label{fig:open-ended-example}
    \vspace{-6mm}
\end{figure}

\subsubsection{Clue-aided Open-Ended QA Evaluation}
Finally, CG-Bench supports open-ended QA evaluation for more comprehensive assessment results. Previous works such as MM-Vet~\citep{yu2023mm-vet} and MMBench-Video~\citep{fang2024mmbenchvideo} use LLMs to evaluate open-ended QA for images and short videos.
In contrast, long videos typically contain more complex information, thus user-generated questions tend to be ambiguous. As a result, the correct answer can be in many distinct forms, which can cause discrepancies between the LLM-evaluated score and the real model's QA ability, as shown in Figure~\ref{fig:open-ended-example}.

To address this, we leverage a low-hallucination MLLM to evaluate the similarity between the text output and the visual information. We choose GPT4o~\citep{gpt4o} as the multimodal evaluator because it ranks among the top in several well-known benchmarks, such as OpenCompass~\citep{2023opencompass}, Lmsys leaderboard~\citep{chiang2024lmsys-leaderboard}, etc., and shows relatively lower hallucinations than other MLLMs. Since directly using GPT-4o for multimodal judging can still introduce hallucination errors and incur high API costs, we propose a heuristic evaluation method to mitigate biases and reduce costs.
First, GPT-4o assesses whether the output can be evaluated based solely on the text answer. If feasible, it outputs either \texttt{yes} or \texttt{no}; otherwise, it requests visual cues by stating ``I need visual clues". This prompts the inclusion of supplementary visual data in the prompt to aid GPT-4o in its evaluation process.
By using pre-annotated time intervals with question clues, we sample frames as visual aids, further reducing hallucination errors and costs. We quantitatively analyze this evaluation method in Sec~\ref{sec:analysis}. More details can be found in the supplementary materials.

\section{Experiments}

In this section, we evaluate a wide range of MLLMs using CG-Bench. We first introduce the evaluation setup, followed by quantitative results for both closed-source and open-source models. Finally, we analyze some key factors in the evaluation.

\subsection{Settings}
We first briefly describe the settings used in our experiments. The supplementary material provides more detailed settings.

\textbf{Models.}
We evaluate the performance of three mainstream commercial models on our CG-Bench: GPT4o~\citep{gpt4o}, Gemini-1.5~\citep{gemini}, and Claude-3.5, including their different versions. 
Also, we assess the representative open-source video models such as LLaVA-OneVision~\citep{li2024llava-onevision}, Qwen2-VL~\citep{Qwen2VL} and InternVL2~\citep{chen2024internvl2}, among others.

\textbf{Frame Sampling.}
For long video understanding, the frame sampling strategy significantly impacts evaluation results.
For open-source MLLMs, we make the best use of our computational resources to use as many frames as possible. For closed-source MLLMs, since the local computational resource is no longer a bottleneck, we can use even more frames. We uniformly sample 128 frames for Long-video MCQ, and use 32 frames as the for Clue-based MCQ.

\textbf{Modality.}
We also explore other modalities: subtitles and audio. For subtitles, we employ a uniform sampling method. If the timestamp of a sampled frame falls within the time interval of a subtitle, that subtitle will be included in the analysis. Each subtitle is considered only once to avoid redundancy.

\textbf{Prompt.}
For MCQ tasks, the model is prompted to provide the uppercase letter corresponding to the correct option. In Open-Ended QA tasks, the model responds freely based on the questions. For the Clue Grounding task, we append the timestamps of each frame and subtitle to enhance the model’s time-awareness, requiring it to return nested lists in the format \texttt{[[s1, e1], [s2, e2], ...]}. 
For open-ended evaluation, we require the model to assess the correctness between the predictions and the ground truth and respond with \texttt{yes} or \texttt{no}.

\begin{table}[t]
\caption{Performance of various open-source and closed-source MLLMs on CG-Bench. We provide human evaluation for showing annotation agreements and the difficulty of our benchmark.
}
\label{tab:main-results}
\centering
\scriptsize
\setlength{\tabcolsep}{1.25mm}{
\begin{tabular}{lrrrrrrrrrr}
\toprule
\textbf{Models}           & \textbf{LLM} & \multicolumn{2}{c}{\textbf{\#F}} & \multicolumn{2}{c}{\textbf{MCQ}} & \multicolumn{4}{c}{\textbf{Cred. Eval.}} &\textbf{OE} \\ 
\cmidrule(lr){3-4} \cmidrule(lr){5-6} \cmidrule(lr){7-10} \cmidrule(lr){11-11}
 &        \textbf{\#param}         & \textbf{clue} & \textbf{long} & \textbf{clue-acc.} & \textbf{long-acc.} & \textbf{mIoU} & \textbf{rec.@IoU} & \textbf{acc.@IoU}  & \textbf{CRR} & \textbf{acc.}  \\ 
\midrule
Random     & -  & - & - & 14.2 &   14.2 & 0.13   &   0.16    &   0.09 &  100  &  0 \\      
Human (full-video) & -  & - & - & 92.2 &     90.3     &   35.5 &  51.2   &  29.8 &  97.9 & 83.7 \\
Human (sparse frames) & -  & - & 128 &     - & 59.9     &   - & -  &  -   &  - & - \\
GPT4o (text)     & -  & 0 & 0 & 16.8 &   16.8 & 0.14  &  0.2   &  0.15   &  100  &  2.1 \\  
\midrule
\multicolumn{11}{c}{\textbf{Open-source MLLMs}} \\  
\midrule
Video-LLAVA~\citep{videollava}               & 7B                 & 8 & 8 & 34.2 & 16.2 & 1.13 & 1.96 & 0.59 & 47.4 & 12.3 \\
VideoLLAMA~\citep{videollama}                & 7B                 & 32 & 32 & 36.8 & 18.4 & 1.21 & 1.87 & 0.84 & 50.0 & 15.8 \\
Videochat2~\citep{mvbench}                & 7B                 & 16 & 16 & 35.2 & 19.3 & 1.28 & 1.98 & 0.94 & 54.8 & 18.6 \\
Qwen-VL-Chat~\citep{Qwen-VL}              & 7B                 & 4 & 4 & 38.3 & 21.6 & 0.89 & 1.19 & 0.42 & 56.4 & 19.4 \\
ST-LLM~\citep{st-llm}                    & 7B                 & 32 & 64 & 39.6 & 23.8 & 2.23 & 2.86 & 1.13 & 60.1 & 20.7 \\
ShareGPT4Video~\citep{chen2024sharegpt4video}            & 16B                & 16 & 16 & 41.4 & 26.7 & 1.85 & 2.65 & 1.01 & 64.5 & 22.0 \\
Chat-UniVi-v1.5~\citep{jin2024chatuniv}            & 13B                & 32 & 64 & 41.5 & 25.9 & 2.07 & 2.53 & 1.21 & 62.4 & 21.4 \\
ViLA~\citep{lin2024vila}          & 8B                 & 14 & 14 & 41.8 & 28.7 & 1.56 & 2.89 & 1.35 & 68.7 & 24.0 \\
GroundVQA~\citep{liu2024etbench} & 0.25B & - & 1200 & 27.3 & - & 1.33 & 1.37 & - & - & - \\
GeLM~\citep{chen2024multihop-egoqa} & 7B & - & 100 & - & - & 2.25 & 2.81 & - & - & - \\
ET-Chat~\citep{liu2024etbench} & 4B & - & 1fps & 17.6 & - & 1.38  & 1.43 & - & - & -\\
InternVL-Chat-v1.5~\citep{chen2023internvl}        & 20B                & 10 & 10 & 42.5 & 28.9 & 2.18 & 2.38 & 1.15 & 68.0 & 23.1 \\
MiniCPM-v2.6~\citep{yao2024minicpm-v}     & 8B  & 32 & 32 & 44.6 & 30.1 & 2.35 & 2.61 & 1.04 & 67.5 & 26.6 \\
LongVA~\citep{zhang2024longva}                    & 7B                 & 32 & 128 & 42.8 & 28.7 & 2.94 & 3.86 & 1.78 & 67.1 & 25.1 \\
LLaVA-OneVision~\citep{li2024llava-onevision}           & 7B                 & 16 & 16 & 43.2 & 31.1 & 1.63 & 1.78 & 1.08 & 72.0 & 25.4 \\
Video-CCAM~\citep{fei2024video-ccam}                & 14B                & 32 & 96 & 43.6 & 29.7 & 2.63 & 3.48 & 1.83 & 68.1 & 25.3 \\
Kangaroo~\citep{liu2024kangaroo}                  & 8B                 & 32 & 64 & 45.9 & 30.2 & 2.56 & 2.81 & 1.94 & 65.8 & 24.5 \\
VITA~\citep{fu2024vita}                      & 8x7B               & 32 & 32 & 47.8 & 33.3 & 3.06 & 3.53 & 2.06 & 69.7 & 27.5 \\
Qwen2-VL~\citep{Qwen2VL}                  & 72B & 32 & 128 & 56.2 & 41.3 & 3.58 & 5.32 & 3.31 & 73.5 & 33.6 \\
InternVL2~\citep{chen2024internvl2}                 & 78B                & 32 & 32 & 58.5 & 42.2 & 3.91 & 5.05 & 2.64 & 72.1 & 32.5 \\
\midrule
\multicolumn{11}{c}{\textbf{Closed-source MLLMs}} \\ 
\midrule
GPT-4o-08-06~\citep{gpt4o}        & -                  & 32 & 128 & 58.3 & 45.2 & 5.62 & 8.30 & 4.38 & 77.5 & 39.5 \\
GPT-4mini-08-06~\citep{gpt4o}        & -                  & 32 & 128 & 48.3 & 33.4 & 3.75 & 5.18 & 2.21 & 69.2 & 25.4 \\
Gemini-1.5-Pro~\citep{gemini}          & -                  & 32 & 128 & 50.1 & 37.2 & 3.95 & 5.81 & 2.53 & 74.3 & 29.3 \\
Gemini-1.5-Flash~\citep{gemini}             & -                  & 32 & 128 & 47.0 & 32.3 & 3.67 & 5.44 & 2.45 & 68.7 & 26.3 \\
Claude3.5-Sonnet             & -                  & 32 & 50 & 56.2 & 40.5 & 3.99 & 5.67 & 2.79 & 72.1 & 35.2 \\
\bottomrule
\end{tabular}%
}
\vspace{-5mm}
\end{table}

\subsection{Main Results}

As shown in Table~\ref{tab:main-results}, the closed-source MLLM GPT4o~\citep{gpt4o} achieved a significant overall lead, surpassing other MLLMs across all metrics. 
Notably, GPT4o's \textbf{long-acc.} approaches 45.2\%, significantly higher than Gemini-1.5-Pro~\citep{gemini}, highlighting its strong capabilities in long video understanding.
For open-source MLLMs, Qwen2-VL's~\citep{Qwen2VL} performance is undeniably impressive, achieving comparable results to GPT4o on \textbf{long-acc.} and \textbf{clue-acc.}.
Other models achieve sub-optimal performance due to the lack of supporting enough context or sufficient training on videos.
Although these MLLMs achieve relatively high accuracy on the MCQ task, they all experienced significant performance degradation when subjected to credibility and open-ended evaluation of CG-Bench. For example, GPT-4o's \textbf{long-acc.} dropped from 45.2 to 4.38 in \textbf{Acc@IoU} and 39.5 in \textbf{OE-acc.}. Notably, with the same number of sampling frames, GPT-4o achieves a \textbf{CRR} of 77.5, while Gemini1.5-Pro only obtains 74.3. This indicates that Gemini-1.5-Pro has an inferior ability to retrieve short-term clues from long videos. Overall, the current MLLMs do not perform well on our CG-Bench, suggesting that there is still considerable room for improvement in their capability and credibility.

Since it is difficult to input more than 128 frames due to the hardware limitations, we alternatively conducted a human evaluation experiment under constrained visual conditions, to see how severe the ``undersampling" issue is for longer video. We uniformly sampled 30 videos from CG-Bench, resulting in 296 questions. For each video, we uniformly sampled 128 frames and asked volunteers to perform an MCQ testing. The resulting accuracy was 59.85\% (row 3 in Table~\ref{tab:main-results}). This result indicates that our dataset is indeed challenging and that it is difficult to derive solutions from a limited number of frames. It also highlights that even the most advanced models, such as GPT-4o, have ample room for improvement in long video comprehension.

\begin{table}[t]
\centering
\scriptsize
\caption{Impact of different prompts and modalities. Each prompt can be composed of frames (F), frame timestamps (FT), subtitles (S), subtitle timestamps (ST), and audio (A). We conduct the main experiments with GPT4o-0806~\citep{gpt4o} while studying the audio modality with Gemini-1.5 Pro~\citep{gemini}.}
\label{tab:impact-prompt-modal}

\setlength{\tabcolsep}{2.0mm}{
\begin{tabular}{llllllll}
\toprule
\textbf{model}  &\textbf{prompt \& modality} &\textbf{clue-acc.} &\textbf{long-acc.} &\textbf{mIoU} &\textbf{Acc@IoU} & \textbf{CRR}  & \textbf{OE-acc.} \\ 
\midrule
GPT4o &S (128 frames) & - & 31.5 & - & - & - & - \\
GPT4o &S (full-video) & - & 34.3 & - & - & - & - \\
\midrule
GPT4o &F & 65.8 & 51.8 & 3.39 & 10.7 & 78.7 & 35.4 \\
GPT4o &F+FT & 65.3$_{(-0.5)}$ & 51.6$_{(-0.2)}$ & 5.73$_{(+2.34)}$ & 20.4$_{(+9.7)}$ & 79.0$_{(-0.3)}$ & 36.8$_{(+1.4)}$ \\
GPT4o &F+S & 66.7$_{(+0.9)}$ & 53.4$_{(+1.6)}$ & 3.96$_{(+0.57)}$ & 11.2$_{(+0.5)}$ & 80.1$_{(+1.4)}$ & 38.2$_{(+2.8)}$ \\
GPT4o &F+S+ST &  67.1$_{(+1.3)}$ & 54.1$_{(+2.3)}$ & 5.19$_{(+1.80)}$ & 13.2$_{(+2.5)}$ & 80.6$_{(+1.9)}$ & 38.4$_{(+3.0)}$ \\
GPT4o &F+S+FT & 67.4$_{(+1.6)}$ & 53.2$_{(+1.4)}$ & 7.80$_{(+4.41)}$ & 22.3$_{(+11.6)}$ & 78.9$_{(+0.2)}$ & 37.9$_{(+2.5)}$ \\
GPT4o &F+S+ST+FT & \textbf{67.5}\textcolor{mygreen}{$_{(+1.7)}$} & \textbf{54.9}\textcolor{mygreen}{$_{(+3.0)}$} & \textbf{9.68}\textcolor{mygreen}{$_{(+6.29)}$} & \textbf{26.7}\textcolor{mygreen}{$_{(+16.0)}$} & \textbf{81.3}\textcolor{mygreen}{$_{(+2.6)}$} & \textbf{39.5}\textcolor{mygreen}{$_{(+4.1)}$}\\
\midrule
Gemini &F+S+ST+FT & 62.1 & 45.1 & 9.16 & 20.7 & 72.6 & 23.2 \\
Gemini &F+S+ST+FT+A & 62.3$_{(+0.2)}$ & 45.0$_{(-0.1)}$ & 9.10$_{(-0.06)}$ & 19.8$_{(-0.9)}$ & 72.2$_{(-0.4)}$ & 23.5$_{(+0.3)}$ \\
\bottomrule
\end{tabular}
}
\vspace{-3mm}
\end{table}

\begin{figure*}[t]
    \centering
    \begin{subfigure}[b]{0.195\textwidth}
        \includegraphics[width=\textwidth]{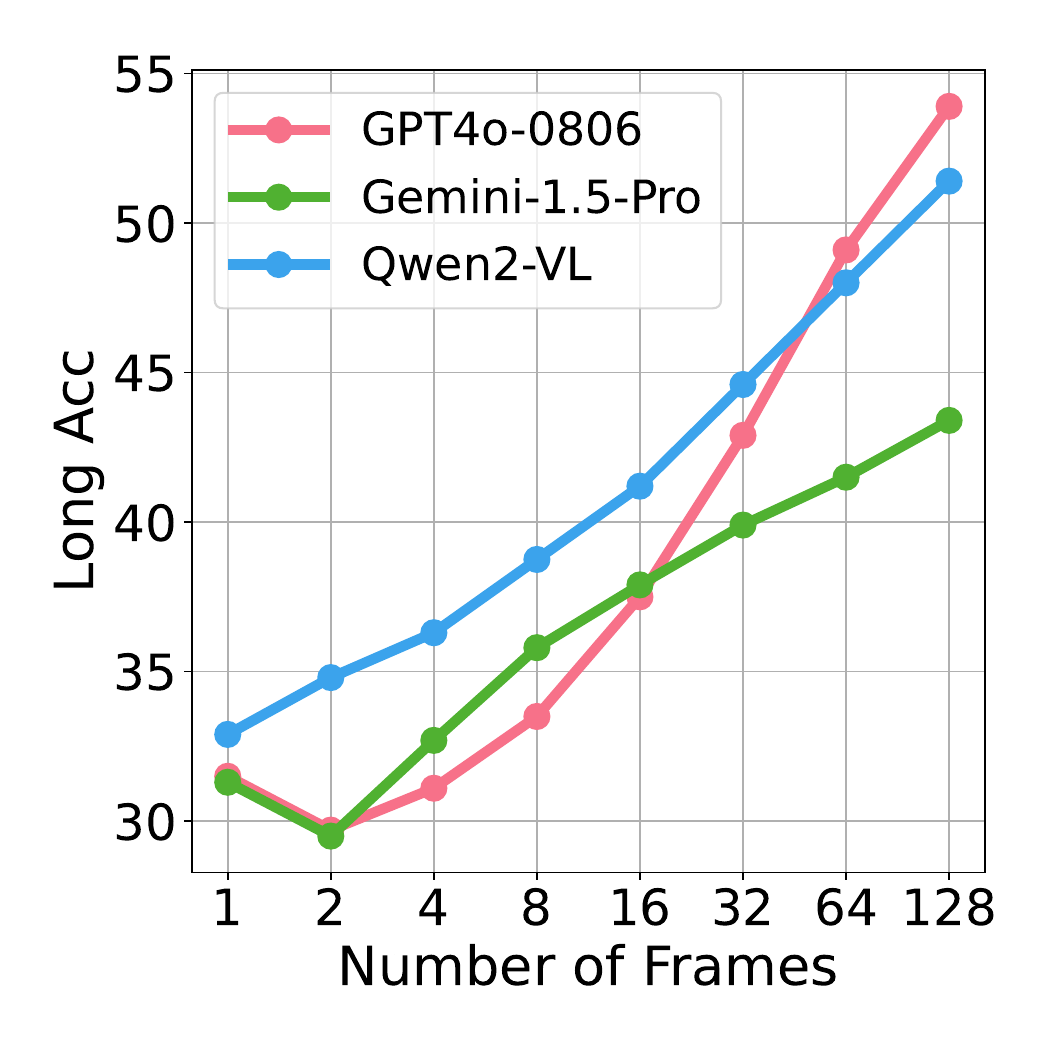}
        \caption{Long Acc}
        \label{fig:subfig1}
    \end{subfigure}
    \hfill
    \begin{subfigure}[b]{0.195\textwidth}
        \includegraphics[width=\textwidth]{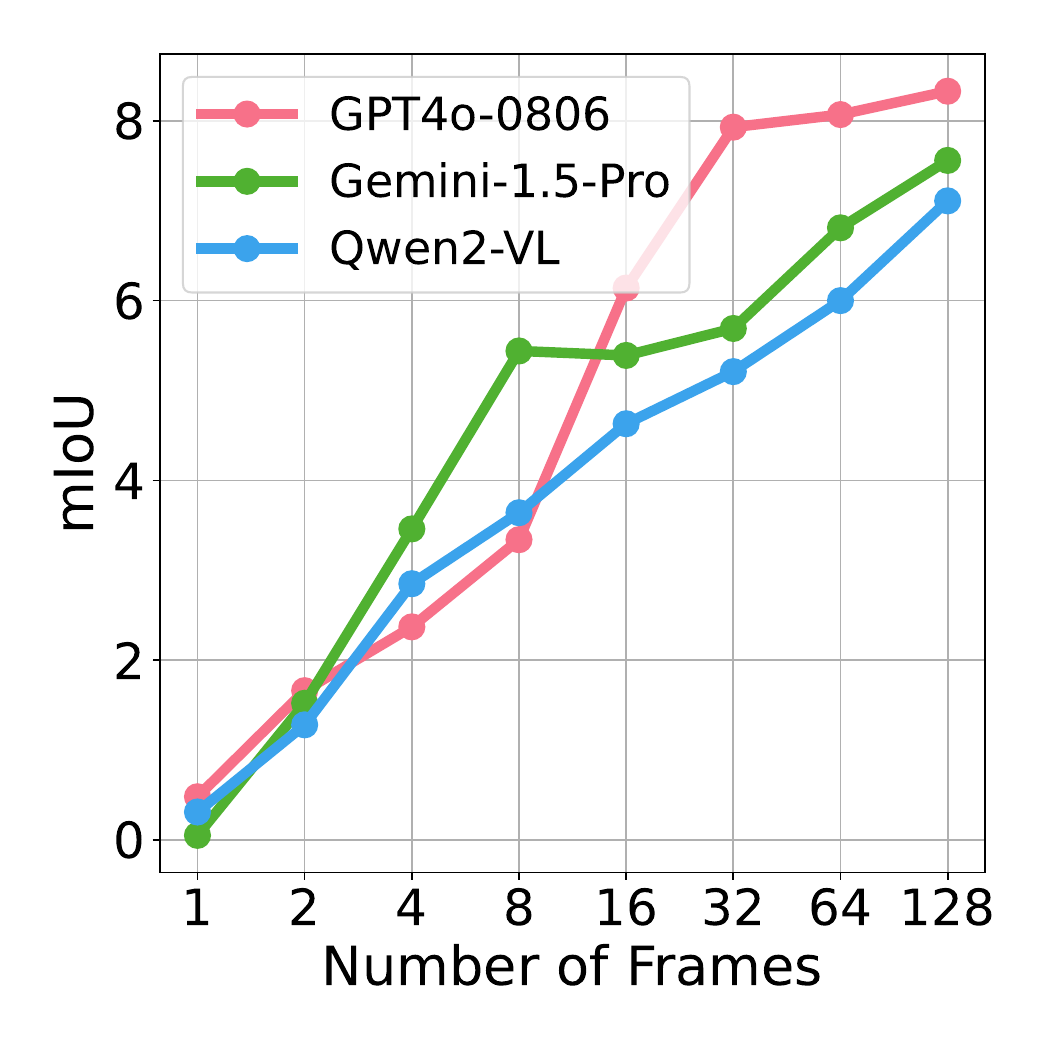}
        \caption{mIoU}
        \label{fig:subfig2}
    \end{subfigure}
    \hfill
    \begin{subfigure}[b]{0.195\textwidth}
        \includegraphics[width=\textwidth]{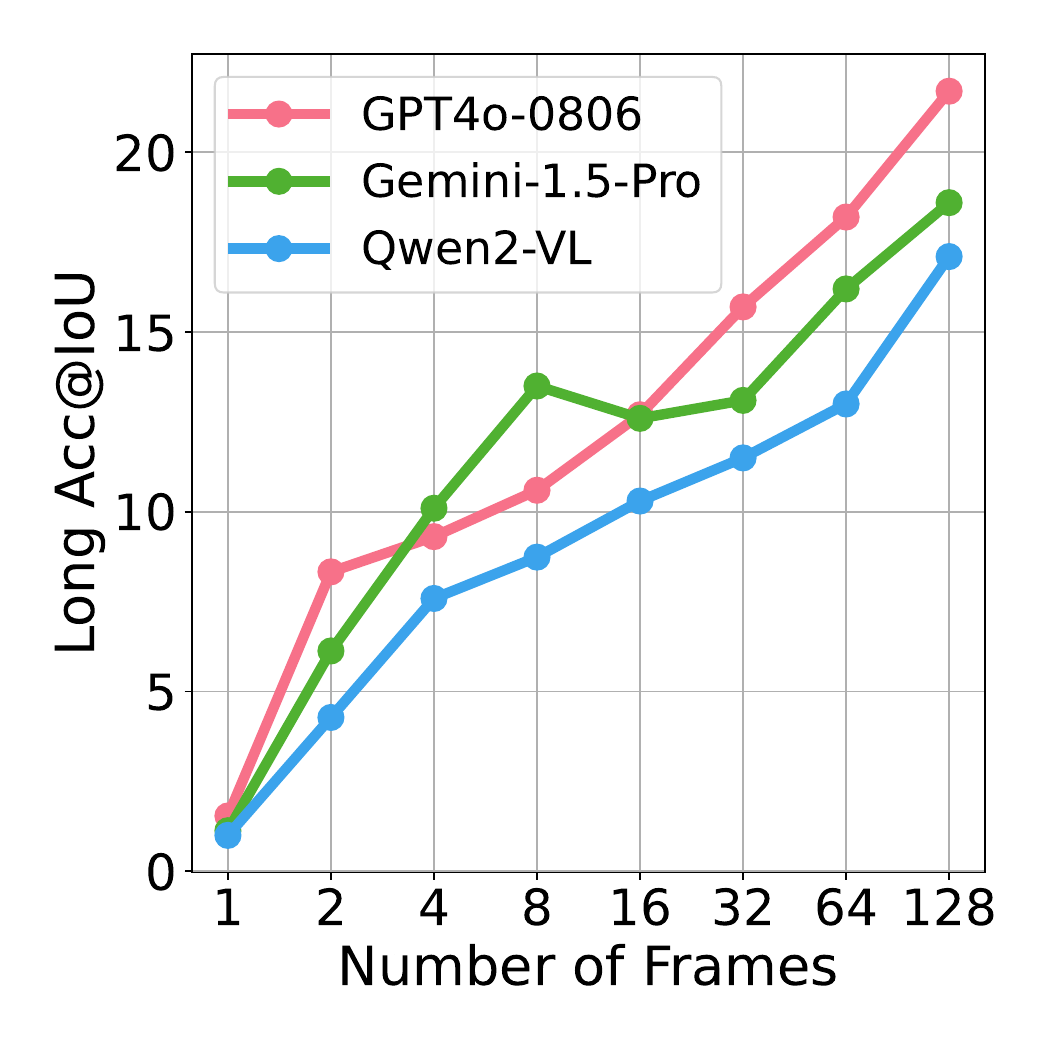}
        \caption{Acc@IoU}
        \label{fig:subfig3}
    \end{subfigure}
        \hfill
    \begin{subfigure}[b]{0.195\textwidth}
        \includegraphics[width=\textwidth]{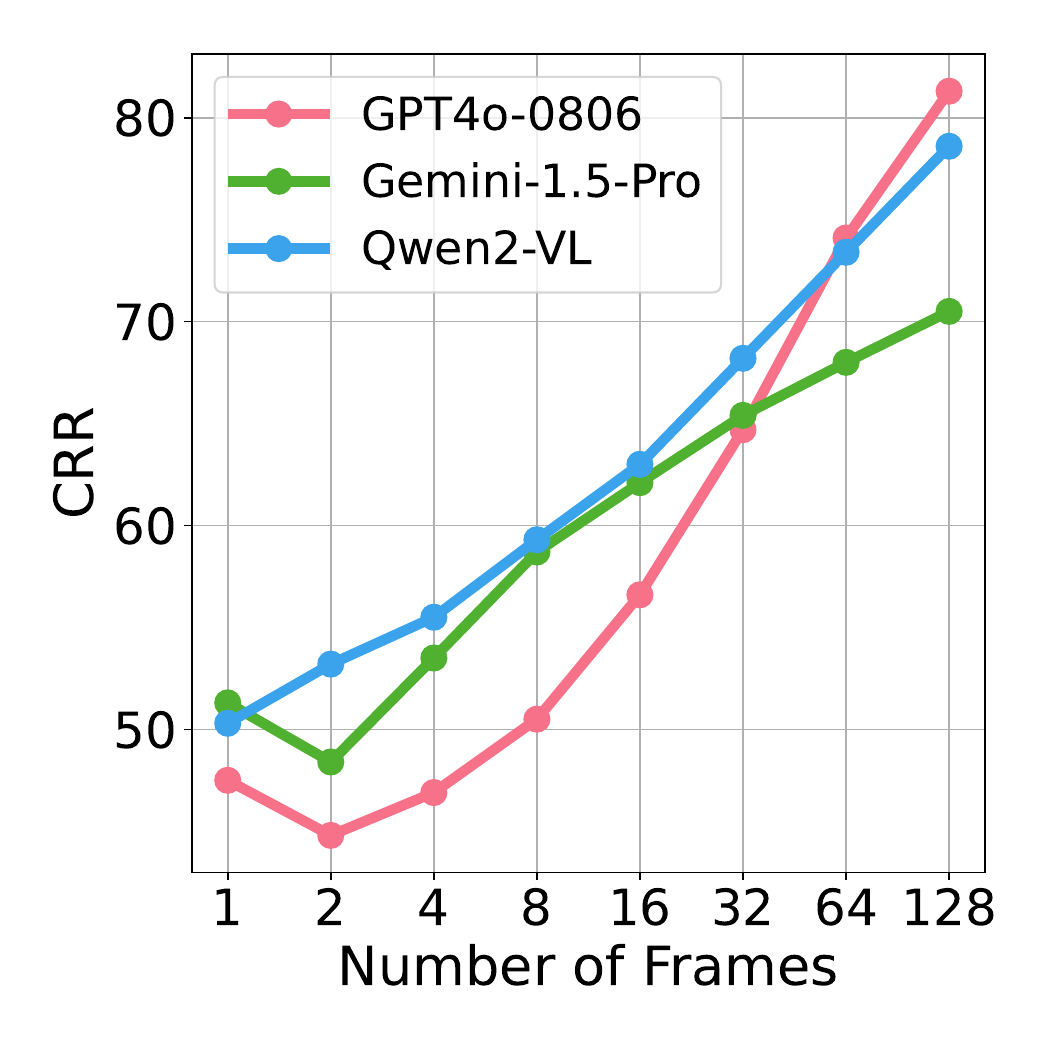}
        \caption{CRR}
        \label{fig:subfig3}
    \end{subfigure}
    \hfill
    \begin{subfigure}[b]{0.195\textwidth}
        \includegraphics[width=\textwidth]{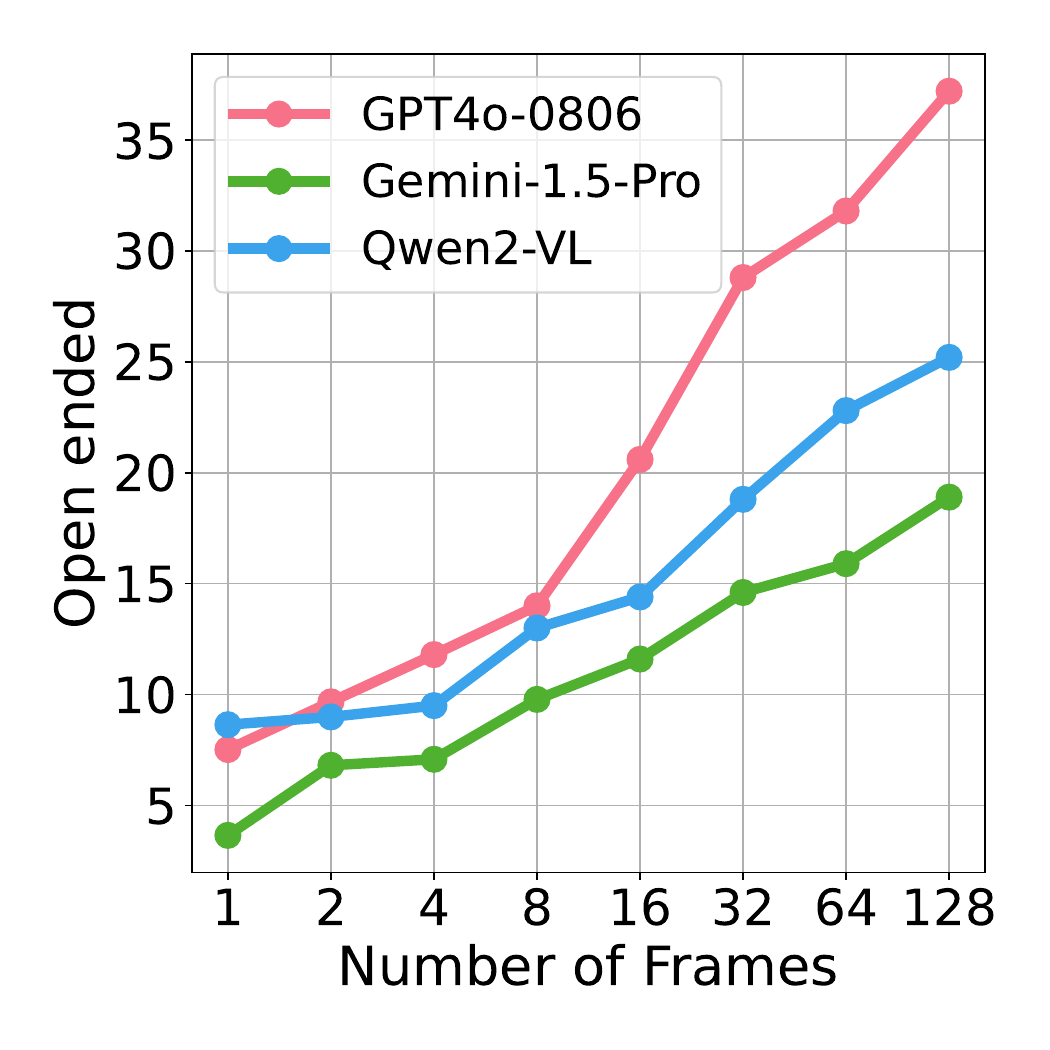}
        \caption{Open ended}
        \label{fig:subfig3}
    \end{subfigure}

    \caption{Impact of sampling frame numbers on different metrics for GPT-4o-0806~\citep{gpt4o}, Gemini-1.5 Pro~\citep{gemini} and Qwen2VL-72B~\citep{Qwen2VL}.}
    \label{fig:impact-num-frames}
    \vspace{-5mm}
\end{figure*}

\subsection{Analysis}
\label{sec:analysis}

Furthermore, we perform a comprehensive analysis of the two leading closed-source MLLMs, GPT4o~\citep{gpt4o} and Gemini-1.5 Pro~\citep{gemini}, as well as the best performing open-source MLLM, Qwen2-VL~\citep{Qwen2VL}, on our CG-Bench. In this analysis, we use 1000 QAC triplets sampled uniformly from all annotations for fast experiments. We report acc.@IoU with $\tau=0$ for a more obvious comparison.

\textbf{Impact of Prompt \& Modality.} As shown in Table~\ref{tab:impact-prompt-modal}, we conduct the ablation studies on the subset that contains subtitles and explore the impact of different prompts on GPT4o and the effect of the audio modality on Gemini-1.5 Pro. Our findings indicate that all prompt types (FT/S/ST), except video frames (F), provide performance benefits across most metrics. Subtitles contribute more to \textbf{long-acc.} than they do to \textbf{clue-acc.}. Additionally, the inclusion of timestamp information (FT/ST) is critical for interval prediction. Timestamps from both frames and subtitles enhance IoU-related metrics, revealing a complementary effect. When both FT and ST are added simultaneously, \textbf{mIoU} increases from 3.39 to 9.68, and \textbf{Acc@IoU} rises from 10.7 to 26.7. When S, FT, and ST are all used in the prompt, the model achieves the best performance across all metrics. In contrast, our exploration of the audio modality (A) revealed that audio does not yield significant performance gain and, in some cases, even slightly degrades the results, as shown in Table~\ref{tab:impact-prompt-modal}. Finally, we conduct experiments using only subtitles from 128 frames versus the full video. The results show that while subtitles offer useful semantic cues, their impact is significantly reduced when visual input is included. This suggests that our benchmark favors visual signals.

\begin{figure*}[t]
    \centering
    \begin{minipage}[b]{0.48\textwidth}
        \centering
        \includegraphics[width=1.0\textwidth]{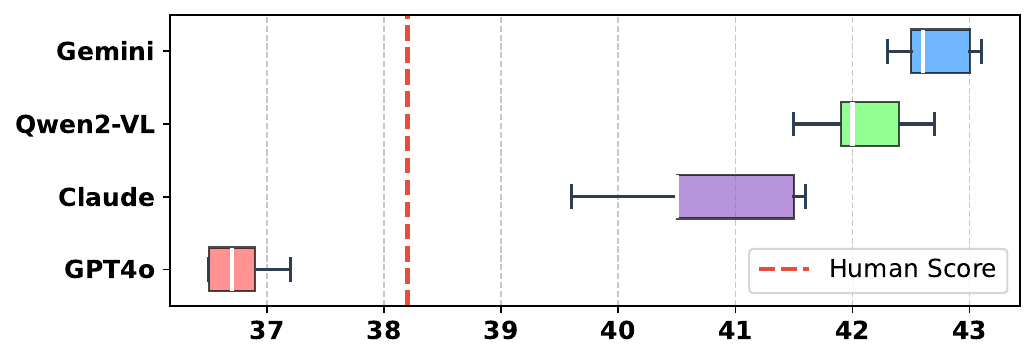}
        \caption{Comparison of using different LLMs as open-ended evaluators for GPT-4o's outputs.}
        \label{fig:impact-open-ended-judge}
    \end{minipage}
    \hfill
    \begin{minipage}[b]{0.48\textwidth}
        \centering
        \scriptsize
        \setlength{\tabcolsep}{1.6mm}{
        \begin{tabular}{lcccc}
            \toprule
            & GT & GT+Vis & Vis & \textbf{Ours} \\
            \midrule
            Bias(\%)$\downarrow$ & 12.4& 6.4&17.0&\textbf{1.0} \\
            Time (s)$\downarrow$ & \textbf{741}& 20,040& 19,640&3,600\\
            Price (\$)$\downarrow$ & \textbf{0.05}& 6.1& 6& 2\\
            \midrule
            Trigger Rate (\%)$\downarrow$ & 0 & 100 & 100 & \textbf{14}  \\
            Trigger Recall Rate (\%)$\uparrow$ & 0 & 100 & 100 & \textbf{88}  \\
            \bottomrule
        \end{tabular}
        }
        \captionof{table}{Comparison of different modes: GT-only, visual-only, GT+vision and heuristic (Ours).}
        \label{tab:open-ended-prompts}
    \end{minipage}
    \vspace{-3mm}
\end{figure*}

\textbf{Impact of Frame Number.} As illustrated in Figure~\ref{fig:impact-num-frames}, we conducted experiments to analyze the performance across various metrics as the number of frames increases. Overall, the performance of all three MLLMs gradually improves with the addition of more frames, with GPT-4o consistently outperforming the others across all metrics. For \textbf{long-acc.} and \textbf{OE acc.}, Qwen2VL achieves performance comparable to GPT-4o. However, compared with Qwen2VL, Gemini excels in terms of mIoU and Acc@IoU. Regarding \textbf{CRR}, GPT-4o demonstrates greater consistency between clue-acc. and long-acc. across more frames, indicating its superior reliability in long video understanding. For open-ended QA, Gemini's higher refusal rate results in a noticeable decline in performance.

\textbf{Open-ended Evaluation Quality.} 
To assess the stability and accuracy of various MLLMs as evaluators, we utilized four models—Gemini, Qwen2VL, Claude, and GPT-4o—each of which evaluated GPT-4o's predictions five times. Human evaluations of GPT-4o's predictions are also conducted for reference. The results, shown in Figure~\ref{fig:impact-open-ended-judge}, indicate that GPT-4o has the highest stability and the smallest deviation from human-assigned scores.
Furthermore, Table~\ref{tab:open-ended-prompts} explores the impact of different evaluation methods. When evaluators were provided only with ground truth (col. ``GT'') or visual information (col. ``Vis''), the scoring bias (absolute difference) between human and model-based evaluation increased. While fully leveraging visual information (col. ``GT+Vis'') improved evaluation accuracy, it also significantly increased the time and cost required. Our proposed heuristic evaluation method achieves the lowest evaluation bias. Additionally, we manually annotated 200 evaluation samples to determine the necessity of visual request triggers. From the bottom block in Table~\ref{tab:open-ended-prompts}, the statistics show that our method achieved a visual request trigger rate (the probability that the model triggers ``visual clues required") of 14\%. The recall rate of this triggering achieves 88\%. This proves that our approach effectively balances cost and performance.

\begin{figure*}[t]
    \centering
    \begin{minipage}[b]{0.48\textwidth}
        \centering
        \includegraphics[width=1.0\textwidth]{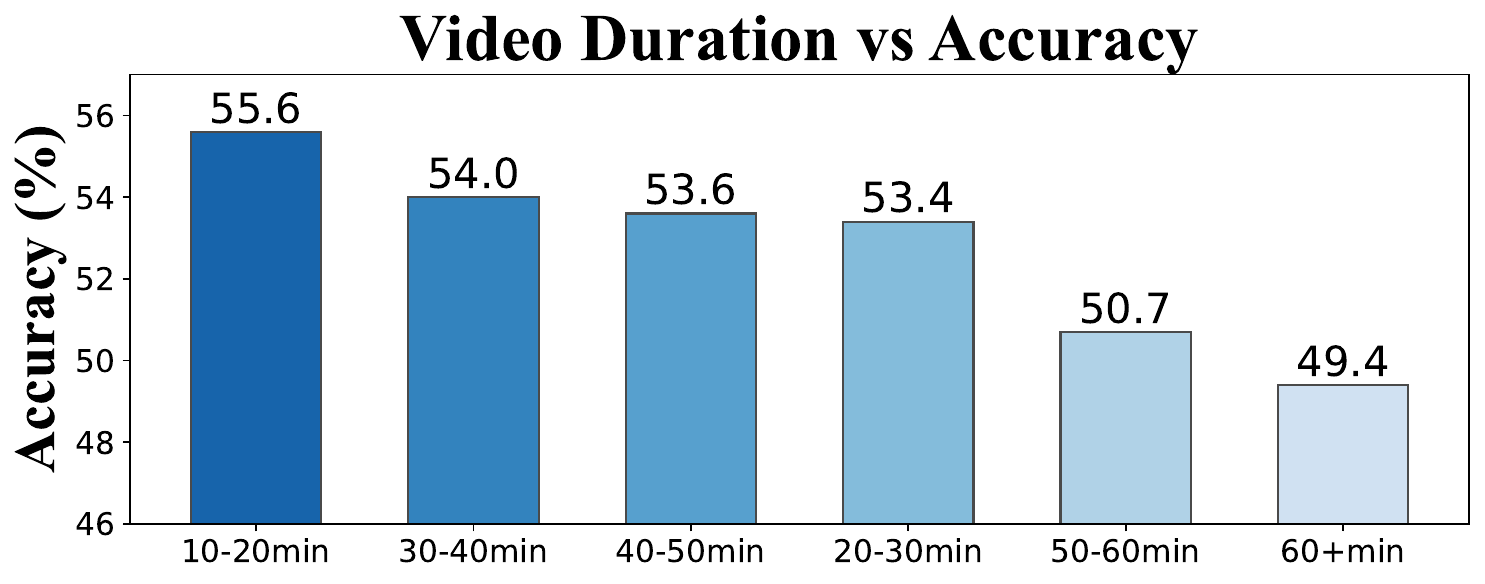}
        \caption{Long-Video-MCQ Accuracy grouped by video duration for GPT4o-0806 with 128 frames. }
        \label{fig:long-acc-group-video-length}
    \end{minipage}
    \hfill
    \begin{minipage}[b]{0.48\textwidth}
        \centering
        \small
        \setlength{\tabcolsep}{0.85mm}{
        \begin{tabular}{cccc}
            \toprule
             \#Frames & Resolution & Sampling Strategy & long-acc. \\
             \midrule
             128 & Low & Uniform & \textbf{53.9} \\
             50 & Low & Uniform & 46.7 \\
             50 & Low & Keyframe & 45.7 \\
             50 & High & Uniform & 51.0 \\
             
            \bottomrule
        \end{tabular}
        }
        \captionof{table}{Impact of different frame sampling strategies on long-acc. for GPT4o-0806.}
        \label{tab:impact-frame-sample}
    \end{minipage}
    \vspace{-6mm}
\end{figure*}

\textbf{Performance grouped by Video Duration.} We grouped videos by duration and evaluated the \textbf{long-acc.} performance of GPT-4o-0806 using 128 frames. Figure~\ref{fig:long-acc-group-video-length} shows that the model struggles with undersampling, especially for longer videos.

\textbf{Impact of Frame Sampling Strategy.} We investigate how different frame sampling strategies affect performance. To expedite testing, we primarily evaluated GPT4o-0806 using 50 uniformly sampled frames, focusing on the \textbf{long-acc} metric. The experiment consists of three parts: 1) low resolution, 2) high resolution, and 3) keyframe extraction (via FFmpeg) combined with low resolution. As shown in Table~\ref{tab:impact-frame-sample}, higher resolution offers some improvement, while keyframe extraction has no significant impact.

\section{Conclusion and Future Work}
In this paper, we introduce CG-Bench, a novel benchmark designed to evaluate clue-grounded question answering capabilities in long video understanding. Unlike existing benchmarks that focus on short videos or rely solely on multiple-choice questions, CG-Bench emphasizes the importance of models retrieving and grounding their answers in specific video segments, enhancing evaluation credibility. By incorporating 1,219 manually curated videos categorized into a detailed three-tier system and 12,129 QA pairs spanning perception, reasoning, and hallucination question types, CG-Bench offers a comprehensive and diverse dataset for assessing MLLMs.
Our two proposed clue-based evaluation methods—clue-grounded white-box and black-box evaluations—provide novel ways to assess whether models genuinely comprehend video content or merely rely on superficial cues. Through extensive experiments involving various closed-source and open-source MLLMs, we found that current models significantly underperform in long video understanding compared to short videos. 
We hope that CG-Bench will serve as a valuable resource for the research community, driving the development of more trustworthy and capable MLLMs for long video understanding.

\bibliography{iclr2025_conference}
\bibliographystyle{iclr2025_conference}

\end{document}